\documentclass[twoside,leqno,twocolumn]{article}

\usepackage[letterpaper]{geometry}

\usepackage{ltexpprt}
\usepackage{hyperref}

\usepackage{times}
\usepackage{epsfig}
\usepackage{graphicx}
\usepackage{amsmath}
\usepackage{amssymb}
\usepackage{subfigure}
\usepackage[table,xcdraw]{xcolor}
\usepackage{multirow}
\usepackage{xcolor}
\usepackage{bm}

\begin{document}

\newcommand\relatedversion{}

\title{\Large DualToken-ViT: Position-aware Efficient Vision Transformer with Dual Token Fusion}
\author{Zhenzhen Chu\thanks{East China Normal University. 51215903091@stu.ecnu.edu.cn}
\and Jiayu Chen\thanks{Alibaba Group. yunji.cjy@alibaba-inc.com}
\and Cen Chen\thanks{East China Normal University. cenchen@dase.ecnu.edu.cn}
\and Chengyu Wang\thanks{Alibaba Group. chengyu.wcy@alibaba-inc.com}
\and Ziheng Wu\thanks{Alibaba Group. zhoulou.wzh@alibaba-inc.com}
\and Jun Huang\thanks{Alibaba Group. huangjun.hj@alibaba-inc.com}
\and Weining Qian\thanks{East China Normal University. wnqian@dase.ecnu.edu.cn}}

\date{}

\maketitle


\fancyfoot[R]{\scriptsize{Copyright \textcopyright\ 2024 by SIAM\\
Unauthorized reproduction of this article is prohibited}}





\begin{abstract} \small\baselineskip=9pt Self-attention-based vision transformers (ViTs) have emerged as a highly competitive architecture in computer vision. Unlike convolutional neural networks (CNNs), ViTs are capable of global information sharing. With the development of various structures of ViTs, ViTs are increasingly advantageous for many vision tasks. However, the quadratic complexity of self-attention renders ViTs computationally intensive, and their lack of inductive biases of locality and translation equivariance demands larger model sizes compared to CNNs to effectively learn visual features. In this paper, we propose a light-weight and efficient vision transformer model called DualToken-ViT that leverages the advantages of CNNs and ViTs. DualToken-ViT effectively fuses the token with local information obtained by convolution-based structure and the token with global information obtained by self-attention-based structure to achieve an efficient attention structure. In addition, we use position-aware global tokens throughout all stages to enrich the global information, which further strengthening the effect of DualToken-ViT. Position-aware global tokens also contain the position information of the image, which makes our model better for vision tasks. We conducted extensive experiments on image classification, object detection and semantic segmentation tasks to demonstrate the effectiveness of DualToken-ViT. On the ImageNet-1K dataset, our models of different scales achieve accuracies of 75.4\% and 79.4\% with only 0.5G and 1.0G FLOPs, respectively, and our model with 1.0G FLOPs outperforms LightViT-T using global tokens by 0.7\%.\end{abstract}

\begin{figure}[h!]
\centering
    \includegraphics[width=\linewidth,trim=203 30 222 12,clip]{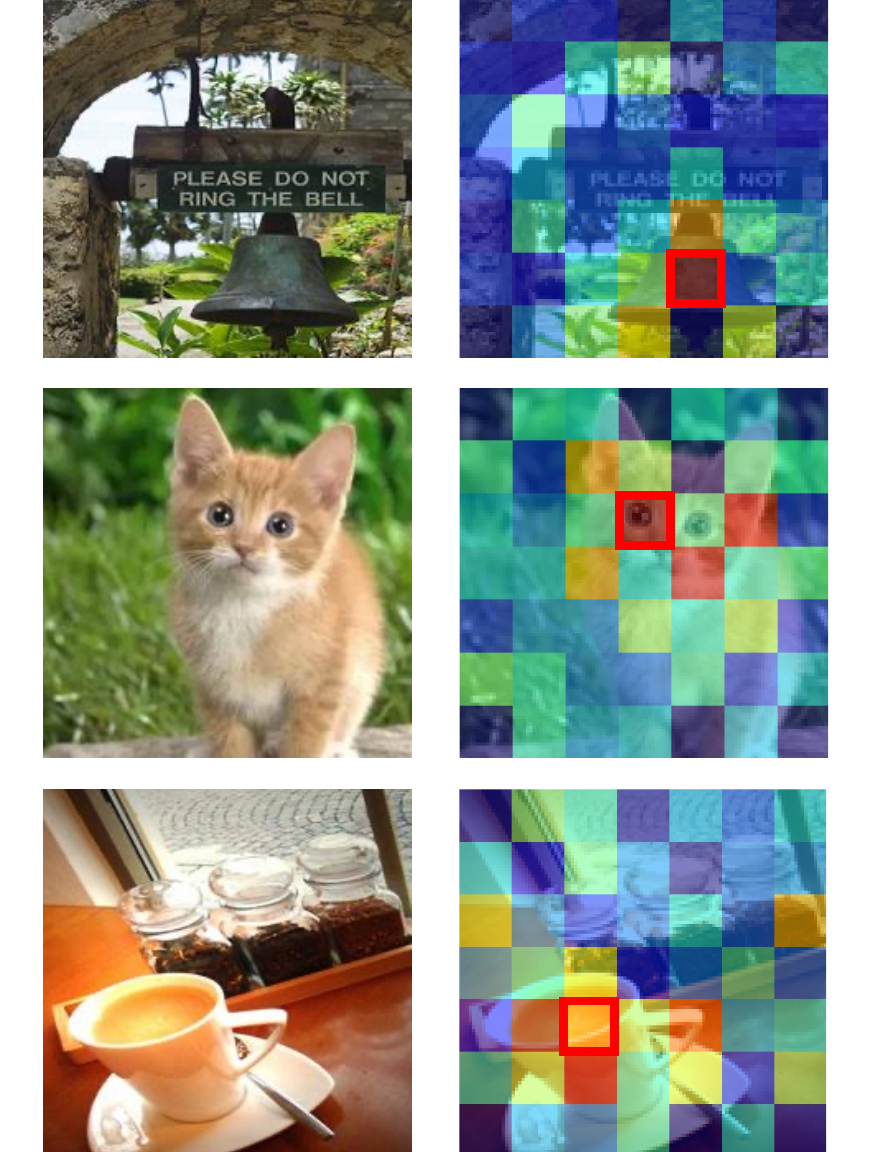}
    \caption{Visualization of the attention map of position-aware global tokens and the key token (the most important part of the image for the image classification task). In each row, the first image is the input of our model, and the second image represents the correlation between the red-boxed portion and each token in the position-aware global tokens containing 7$\times$7 tokens, where the red-boxed portion is the key token of the first image.}
\label{fig:visual_part}
\end{figure}

\section{Introduction}

In recent years, vision transformers (ViTs)
have emerged as a powerful architecture for various vision tasks such as image classification~\cite{dosovitskiy2020vit} and object
detection~\cite{carion2020end, zhu2020deformable}.
This is due to the ability of self-attention to capture global information from the image, providing sufficient and useful visual features, while convolutional neural networks (CNNs) are limited by the size of convolutional kernel and can only extract local information.
As the model size of ViTs and the dataset size increase, there is still no sign of a saturation in performance, which is an advantage that CNNs do not have for large models as well as for large datasets~\cite{dosovitskiy2020vit}. 
However, CNNs are more advantageous than ViTs in light-weight models due to certain inductive biases that ViTs lack. Since the quadratic complexity of self-attention, the computational cost of ViTs can also be high. Therefore, it is challenging to design light-weight-based efficient ViTs.

To design more efficient and light-weight ViTs, \cite{wang2021pvtv1, chu2021twinssvt} propose a pyramid structure that divides the model into several stages, with the number of tokens decreasing and the number of channels increasing by stage. \cite{mehta2022separable, bolya2023hydra} focus on reducing the quadratic complexity of self-attention by simplifying and improving the structure of self-attention, 
but they sacrifice the effectiveness of attention.
Reducing the number of tokens involved in self-attention is also a common approach, e.g.,
\cite{wang2021pvtv1, wang2022pvtv2, pan2022litv2}
downsample the key and the value in self-attention.
Some works~\cite{liu2021swin, dong2022cswin} based on locally-grouped self-attention reduce the complexity of the overall attention part by performing self-attention on grouped tokens separately, but such methods may damage the sharing of global information. 
Some works also add a few additional learnable parameters to enrich the global information of the backbone, for example, \cite{huang2022lightvit, chen2022mobileformer, yao2022dualvit} add the branch of global tokens that throughout all stages. This method can supplement global information for local attention (such as locally-grouped self-attention based and convolution-based structures). These existing methods using global tokens, however, consider only global information and ignore positional information that is very useful for vision tasks.
      
In this paper, we propose a light-weight and efficient vision transformer model called DualToken-ViT. 
Our proposed model features a more efficient attention structure designed to replace self-attention.
We combine the advantages of convolution and self-attention, leveraging them to extract local and global information respectively, and then fuse the outputs of both to achieve an efficient attention structure. Although window self-attention~\cite{liu2021swin} is also able to extract local information, we observe that
it is less efficient than the convolution on our light-weight model. 
To reduce the computational complexity of self-attention in global information broadcasting, we downsample the feature map that produces key and value by step-wise downsampling, which can retain more information during the downsampling process. 
Moreover, we use position-aware global tokens throughout all stages to further enrich the global information. 
In contrast to the normal global tokens~\cite{huang2022lightvit, chen2022mobileformer, yao2022dualvit}, our position-aware global tokens are also able to retain position information of the image and pass it on, which can give our model an advantage in vision tasks. 
As shown in Figure~\ref{fig:visual_part}, the key token in the image generates higher correlation with the corresponding tokens in the position-aware global tokens, which demonstrates the effectiveness of our position-aware global tokens.
In summary, our contributions are as follows:
\begin{itemize}
    \item We design a light-weight and efficient vision transformer model called DualToken-ViT, which combines the advantages of convolution and self-attention to achieve an efficient attention structure by fusing
    local and global tokens containing local and global information, respectively.
    \item
    We further propose position-aware global tokens that contain the position information of the image to enrich the global information.
    \item Among vision models of the same FLOPs magnitude, our DualToken-ViT shows the best performance on the tasks of image classification, object detection and semantic segmentation.
\end{itemize}

\section{Related work}

\noindent {\normalsize \bf Efficient Vision Transformers.} ViTs are first proposed by~\cite{dosovitskiy2020vit}, which applies transformer-based structures to computer vision.~\cite{wang2021pvtv1, chu2021twinssvt} apply the pyramid structure to ViTs, which will incrementally transform the spatial information into the rich semantic information. To achieve efficient ViTs, some works are beginning to find suitable alternatives to self-attention in computer vision tasks, such as~\cite{mehta2022separable, bolya2023hydra}, which make the model smaller by reducing the complexity of self-attention.
\cite{wang2021pvtv1, wang2022pvtv2, pan2022litv2}
reduce the required computational resources by reducing the number of tokens involved in self-attention.~\cite{liu2021swin, dong2022cswin} use locally-grouped self-attention based methods to reduce the complexity of the overall attention part. There are also some works that combine convolution into ViTs, for example,~\cite{wang2022pvtv2, huang2022svit} use convolution-based FFN (feed-forward neural network) to replace the normal
FFN,~\cite{pan2022litv2} uses
more convolution-based structure in the shallow stages of the model and more transformer-based structure in the deep stages of the model.
Moreover, there are also many works that use local information extracted by convolution or window self-attention to compensate for the shortcomings of ViTs, such as~\cite{mehta2021mobilevitv1, pan2022edgevit}.

\begin{figure*}[htbp]
\centering
    \includegraphics[width=0.97\linewidth,trim=6 118 6 95,clip]{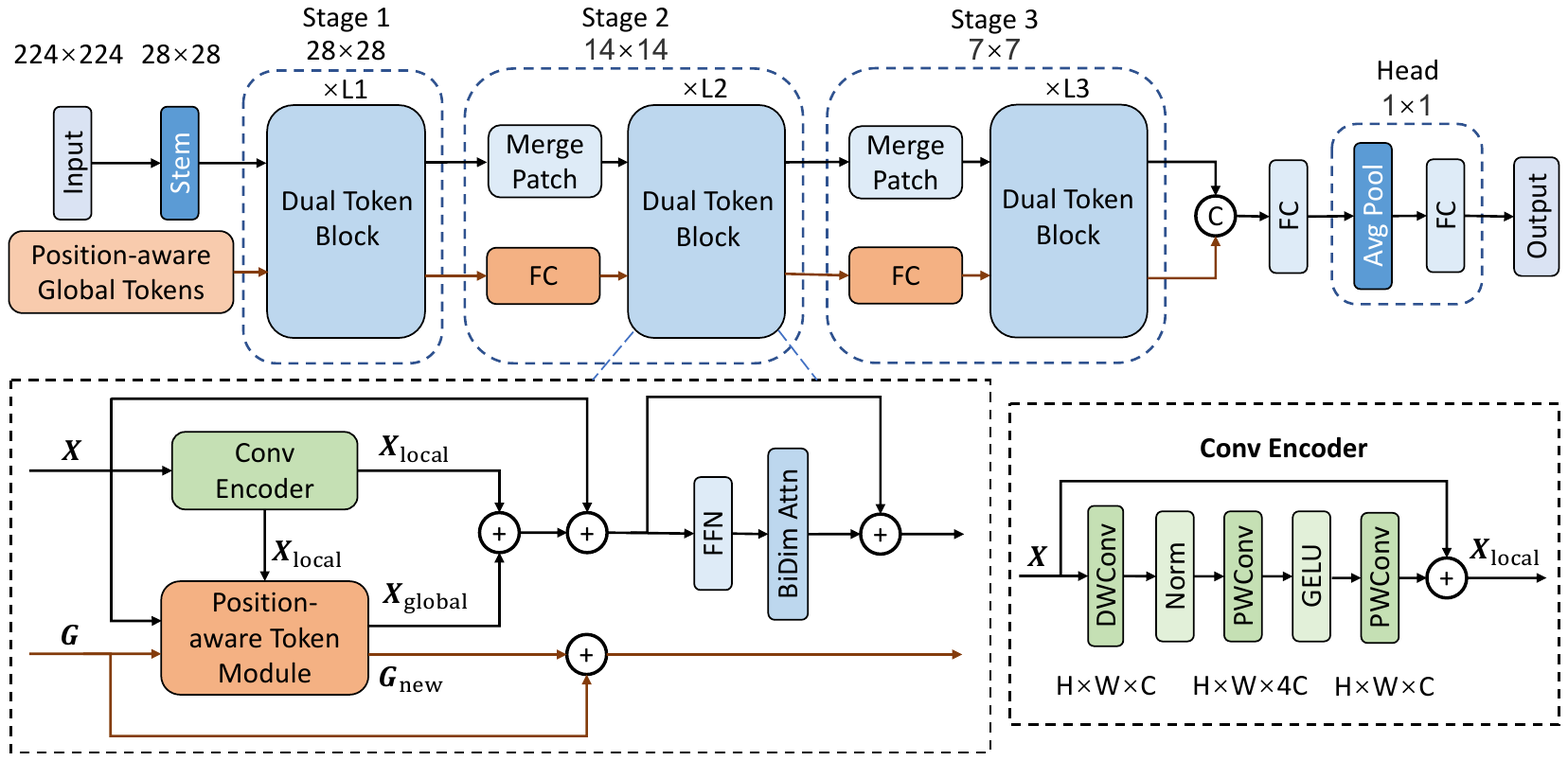}
    \caption{The architecture of DualToken-ViT. \textcircled{+} represents element-wise addition. \textcircled{c} represents concatenation in the token axis.}
\label{fig:DualToken-ViT}
\end{figure*}

\noindent {\normalsize \bf Efficient Attention Structures.} For local attention, convolution works well for extracting local information
in vision tasks,
e.g.,~\cite{mehta2021mobilevitv1, pan2022edgevit}
add convolution to model to aggregate local information. Among transformer-based structures, locally-grouped self-attention~\cite{liu2021swin, dong2022cswin} can also achieve local attention by adjusting the window size, and their complexity will be much less than that of self-attention. For global attention, self-attention~\cite{vaswani2017attention} has a strong ability to extract global information, but on light-weight models, it may not be able to extract visual features well due to the lack of model size. Methods~\cite{huang2022lightvit, chen2022mobileformer, yao2022dualvit} using global tokens can also aggregate global information. They use self-attention to update global tokens and broadcast global information.
Since the number of tokens in global tokens will not be set very large, the complexity will not be very high.
Some works~\cite{pan2022edgevit, huang2022lightvit, pan2022litv2, chen2022mobileformer} achieve a more efficient attention structure by combining both local and global attention. In this paper, we implement an efficient attention structure by combining convolution-based local attention and self-attention-based global attention, and use another branch of position-aware global tokens for the delivery of global and position information throughout the model, where position-aware global tokens are an improvement over global tokens~\cite{huang2022lightvit, chen2022mobileformer, yao2022dualvit}.

\section{Methodology}

As shown in Figure~\ref{fig:DualToken-ViT}, DualToken-ViT is designed based on the 3-stage structure of LightViT~\cite{huang2022lightvit}.
The structure of stem and merge patch block in our model is the same as the corresponding part in LightViT.
FC refers to fully connected layer. There are two branches in our model: image tokens and position-aware global tokens. The branch of image tokens is responsible for obtaining various information from position-aware global tokens, and the branch of position-aware global tokens is responsible for updating position-aware global tokens through the branch of image tokens and passing it on. In the attention part of each Dual Token Block, we obtain information from the position-aware global tokens and fuse local and global information. We also add BiDim Attn (bi-dimensional attention) proposed in LightViT after the FFN. In this section, we mainly introduce two important parts:
the fusion of local and global information and position-aware global tokens.

\begin{figure*}[htbp]
\centering
\subfigure[]{
\includegraphics[width=115pt,trim=333 215 200 150,clip]{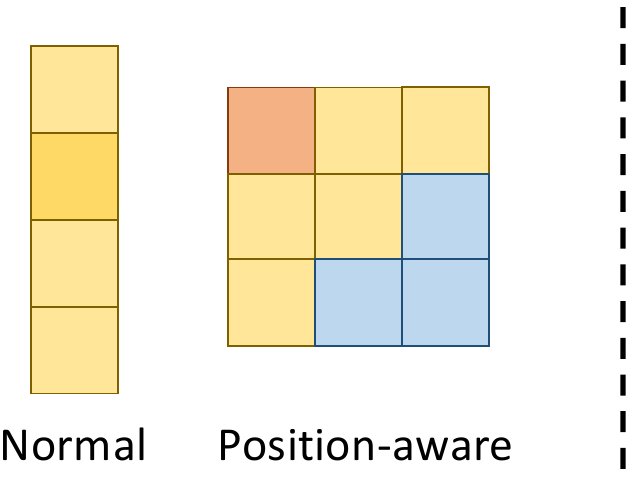}
\label{fig:gtokens}
}
\subfigure[]{
\includegraphics[width=305pt,trim=135 230 150 215,clip]{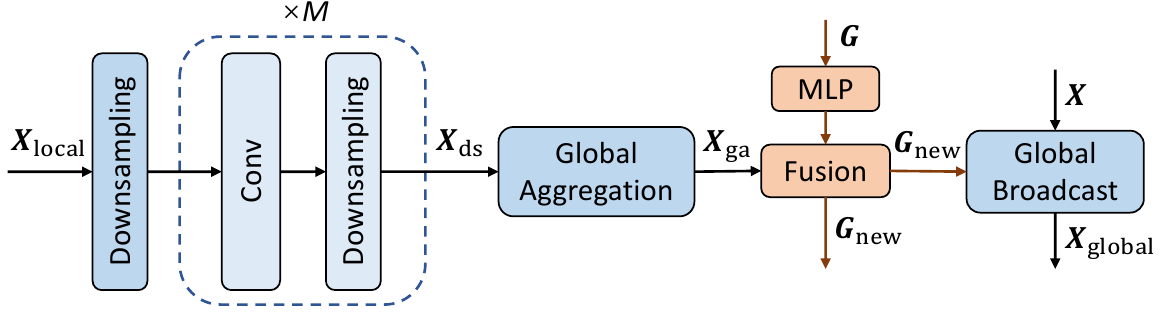}
\label{fig:Position-awareTokenModule}
}
\subfigure[]{
\includegraphics[width=130pt,trim=60 308 585 115,clip]{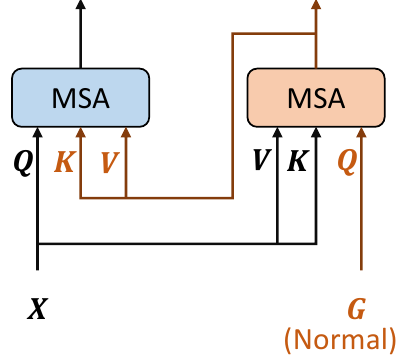}
\label{fig:GlobalToken}
}
\subfigure[]{
\includegraphics[width=130pt,trim=310 307 330 112,clip]{globaltokens.pdf}
\label{fig:PosNormGlobalToken}
}
\subfigure[]{
\includegraphics[width=130pt,trim=550 303 85 115,clip]{globaltokens.pdf}
\label{fig:PosGlobalToken}
}
\caption{(a) shows normal global tokens and position-aware global tokens. (b) shows the structure of Position-aware Token Module using position-aware global tokens. (c), (d) and (e) show different methods of applying global tokens and show only the interaction of $\bm{X}$ and $\bm{G}$, omitting the processing of $\bm{X}$ and $\bm{G}$. MSA represents multi-head self-attention.}
\label{fig:GlobalTokens}
\end{figure*}

\subsection{Fusion of Local and Global Information}

In the attention part of each Dual Token Block, we extract the local and global information through two branches, Conv Encoder (convolution encoder) and Position-aware Token Module, respectively, and then fuse these two parts.

\noindent {\normalsize \bf Local Attention.} We use Conv Encoder for local information extraction in each block of our model, since for light-weight models, local information extraction with convolution will perform better than window self-attention. Conv Encoder has the same structure as the ConvNeXt block~\cite{liu2022convnet}, which is represented as follows:
\begin{equation}
\bm{X}_{\text{local}}=\bm{X}+\mathit{\text{PW}}_2(\text{GELU}(\mathit{\text{PW}}_1(\text{LN}(\text{DW}(\bm{X})))))
\end{equation}
where $\bm{X}$ is the input image tokens of size H$\times$W$\times$C, DW is the depth-wise convolution, $\mathit{\text{PW}}_1$ and $\mathit{\text{PW}}_2$ are point-wise convolution, LN is the layer norm, and $\bm{X}_{\text{local}}$ containing local information is the output of Conv Encoder.

\noindent {\normalsize \bf Position-aware Token Module.} This module is responsible for extracting global information, and its structure is shown in Figure~\ref{fig:Position-awareTokenModule}. In order to reduce the complexity of extracting global information, we first downsample $\bm{X}_{\text{local}}$ containing local information and aggregate the global information. Position-aware global tokens are then used to enrich global information. We end up broadcasting this global information to image tokens. The detailed process is as follows:

(1) Downsampling. If the size of $\bm{X}_{\text{local}}$ is large and does not match the expected size, then it is downsampled twice first. After that, local information is extracted by convolution and downsampled twice, and the process is repeated $M$ times until the feature map size reaches the expected size. Compared with the one-step downsampling method, this step-wise downsampling method can reduce the loss of information during the downsampling process and retain more useful information. The entire step-wise downsampling process is represented as follows:
\begin{equation}
\bm{X}_{\text{ds}} = \phi(\text{DS}(\bm{X}_{\text{local}}))
\end{equation}
where DS represents twice the downsampling using average pooling, $\phi$ represents that if the feature map size does not match the expected size, then several convolution and downsampling operations are performed, with each operation represented by DS(Conv(·)), and $\bm{X}_{\text{ds}}$ represents the result after step-wise downsampling.

(2) Global Aggregation. Aggregation of global information using multi-head self-attention for the $\bm{X}_{\text{ds}}$ output in the previous step:
\begin{equation}
\bm{X}_{\text{ga}} = \text{MSA}(\bm{Q}_{\text{ds}}, \bm{K}_{\text{ds}}, \bm{V}_{\text{ds}})
\end{equation}
where $\bm{Q}_{\text{ds}}$, $\bm{K}_{\text{ds}}$ and $\bm{V}_{\text{ds}}$ are produced by $\bm{X}_{\text{ds}}$ through linear projection, and then $\bm{X}_{\text{ga}}$ containing global information is obtained.

(3) Enrich the global information. Use position-aware global tokens $\bm{G}$ to enrich $\bm{X}_{\text{ga}}$'s global information:
\begin{equation}
\bm{G}_{\text{new}} = \text{Fuse}(\bm{G}, \bm{X}_{\text{ga}})
\end{equation}
where Fuse is how the two are fused, which will be explained later along with position-aware global tokens.

(4) Global Broadcast. The global information in $\bm{G}_{\text{new}}$ is broadcast to the image tokens using self-attention. This process is represented as follows:
\begin{equation}
\bm{X}_{\text{global}} = \text{MSA}(\bm{Q}_{\text{image}}, \bm{K}_\text{g}, \bm{V}_\text{g})
\end{equation}
where $\bm{Q}_{\text{image}}$ is produced by image tokens through linear projection, $\bm{K}_\text{g}$ and $\bm{V}_\text{g}$ are produced by $\bm{G}_{\text{new}}$ through linear projection.

\noindent {\normalsize \bf Fusion.} Fusing the two tokens, which contain local and global information respectively:
\begin{equation}
\bm{X}_{\text{new}} = \bm{X}_{\text{local}} + \bm{X}_{\text{global}}
\end{equation}

\subsection{Position-aware Global Tokens}

Global Aggregation is able to extract global information, but its scope is only in a block.
For this reason,
we employ the position-aware global tokens $\bm{G}$, which throughout all stages, to fuse with the $\bm{X}_{\text{ga}}$ to obtain $\bm{G}_{\text{new}}$. $\bm{G}_{\text{new}}$ has richer global information and can be used to enrich the global information and function as new position-aware global tokens to the next block after adding the identical mapping. In addition to global information, position information in position-aware global tokens is also delivered.

\noindent {\normalsize \bf Global Tokens with Position Information.} Figure~\ref{fig:gtokens} shows the normal global tokens~\cite{huang2022lightvit, chen2022mobileformer, yao2022dualvit} and our position-aware global tokens. The one-dimensional global tokens contain global information, and our two-dimensional position-aware global tokens additionally contain location information. The normal global tokens use the way in Figure~\ref{fig:GlobalToken} to fuse $\bm{X}$ and $\bm{G}$ via multi-head self-attention and broadcast the global information.
Figure~\ref{fig:PosGlobalToken} is our Position-aware Global Tokens, which we set to the same number of tokens as in $\bm{X}_{\text{ga}}$, and use weighted summation to fuse them:
\begin{equation}
\bm{G}_{\text{new}} = \text{Fuse}(\bm{G}, \bm{X}_{\text{ga}}) = \alpha \text{MLP}(\bm{G}) + (1 - \alpha)\bm{X}_{\text{ga}}
\end{equation}
where $\alpha \in [0, 1]$ is a weight that is set in advance. Although our position-aware global tokens will cause the parameters to increase due to the increase in the number of tokens, it will perform better than the normal global tokens.

\noindent {\normalsize \bf MLP.} Before fusion, we use MLP for position-aware global tokens, which allows for a better fusion of the two. The formula of MLP is as follows:
\begin{equation}
\bm{G}'=(\text{Linear}(\text{GELU}(\text{Linear}(\bm{G})))
\end{equation}
Since the normal MLP is only in the channel dimension, we also attempt to use token-mixing MLP~\cite{tolstikhin2021mlp} to additionally extract the information in the spatial dimension:
\begin{equation}
\bm{G}'=\text{Transpose}(\text{Linear}(\text{Transpose}(\text{Linear}(\bm{G}))))
\end{equation}
where Transpose represents the transposition of spatial and channel axis. We refer to this process as MixMLP.

\begin{table}[]
\centering
\caption{Macro structures of two DualToken-ViT variants. B, C and H represent the number of blocks, channels and attention heads in multi-head self-attention, respectively.}
\setlength\tabcolsep{4pt}
{\normalsize
\begin{tabular}{c|c|c|c}
\hline
\textbf{Stage}       & \textbf{Stride} & \textbf{DualToken-ViT-T} & \textbf{DualToken-ViT-S} \\ \hline
Stage 1 & 1/8             & B=2 C=48 H=2             & B=2 C=64 H=2             \\
Stage 2 & 1/16            & B=6 C=96 H=4             & B=6 C=128 H=4            \\
Stage 3 & 1/32            & B=4 C=192 H=8            & B=6 C=256 H=8            \\ \hline
\end{tabular}
}
\label{table:architectures}
\end{table}

\subsection{Architectures}

We design two DualToken-ViT models of different scales, and their macro structures are shown in Table~\ref{table:architectures}. For the task of image classification on the ImageNet-1k~\cite{deng2009imagenet} dataset, we default the size of the image after data augment is 224$\times$224. To prevent the complexity of the model from being too large, we set the size of position-aware global tokens to 7$\times$7. In this way, the $M$ of the first two stages are set to 1 and 0 respectively, and the size of $\bm{X}_{\text{ga}}$ is exactly 7$\times$7. In the third step, the feature map size of the image tokens is exactly 7$\times$7, this eliminates the need for local information extraction and downsampling, and allows these steps to be skipped directly. Furthermore, the convolutional kernel size of depth-wise convolution in the Conv Encoder of the first two stages is 5$\times$5 and 7$\times$7 respectively, and the convolutional kernel sizes in the step-wise downsampling are all 3$\times$3. In addition, if the size of the input image changes (as in the object detection and semantic segmentation tasks) and it is not possible to make $\bm{X}_{\text{ga}}$ the same size as the position-aware global tokens, we use interpolation to change the size of $\bm{X}_{\text{ga}}$ to the same size as the position-aware global tokens. In the fusion of $\bm{G}$ and $\bm{X}_{\text{ga}}$, we set $\alpha$ to 0.1.

\begin{table}[htbp]
\centering
\caption{Image classification performance on ImageNet-1k. ``mix'' indicates that our model uses MixMLP instead of normal MLP.}
\begin{tabular}{c|ccc}
\hline
{\color[HTML]{333333} \textbf{Model}} & \textbf{\begin{tabular}[c]{@{}c@{}}FLOPs\\ (G)\end{tabular}} & \textbf{\begin{tabular}[c]{@{}c@{}}Params\\ (M)\end{tabular}} & \textbf{\begin{tabular}[c]{@{}c@{}}Top-1\\ (\%)\end{tabular}} \\ \hline
MobileNetV2 (1.4)~\cite{sandler2018mobilenetv2}                      & 0.6                                                          & 6.9                                                           & 74.7                                                          \\
MobileViTv1-XXS~\cite{mehta2021mobilevitv1}                       & 0.4                                                          & 1.3                                                           & 69.0                                                          \\
MobileViTv2-0.5~\cite{mehta2022separable}                       & 0.5                                                          & 1.4                                                           & 70.2                                                          \\
PVTv2-B0~\cite{wang2022pvtv2}                             & 0.6                                                          & 3.4                                                           & 70.5                                                          \\
EdgeViT-XXS~\cite{pan2022edgevit}                           & 0.6                                                          & 4.1                                                           & 74.4                                                          \\
\textbf{DualToken-ViT-T (mix)}        & \textbf{0.5}                                                 & \textbf{5.8}                                                  & \textbf{75.4}                                                 \\ \hline
RegNetY-800M~\cite{radosavovic2020designing}                          & 0.8                                                          & 6.3                                                           & 76.3                                                          \\
DeiT-Ti~\cite{touvron2021training}                               & 1.3                                                          & 5.7                                                           & 72.2                                                          \\
T2T-ViT-7~\cite{yuan2021tokens}                             & 1.1                                                          & 4.3                                                           & 71.7                                                          \\
SimViT-Micro~\cite{li2022simvit}                          & 0.7                                                          & 3.3                                                           & 71.1                                                          \\
MobileViTv1-XS~\cite{mehta2021mobilevitv1}                         & 1.0                                                          & 2.3                                                           & 74.8                                                          \\
TNT-Ti~\cite{han2021transformer}                                 & 1.4                                                          & 6.1                                                           & 73.9                                                          \\
LVT~\cite{yang2022lite}                                   & 0.9                                                          & 5.5                                                           & 74.8                                                          \\
EdgeViT-XS~\cite{pan2022edgevit}                            & 1.1                                                          & 6.7                                                           & 77.5                                                          \\
XCiT-T12~\cite{ali2021xcit}                              & 1.3                                                          & 6.7                                                           & 77.1                                                          \\
LightViT-T~\cite{huang2022lightvit}                            & 0.7                                                          & 9.4                                                           & 78.7                                                          \\
DualToken-ViT-S (mix)                 & 1.0                                                          & 11.4                                                          & 79.4                                                          \\
\textbf{DualToken-ViT-S}              & \textbf{1.1}                                                 & \textbf{11.9}                                                 & \textbf{79.5}                                                 \\ \hline
\end{tabular}
\label{table:cls}
\end{table}

\begin{table*}[htbp!]
\centering
\caption{Object detection and instance segmentation performance by Mask R-CNN on MS-COCO. All the models are pretrained on ImageNet-1K.}
\setlength\tabcolsep{3.2pt}
{\small
\begin{tabular}{c|cc|cccccc|cccccc}
\hline
\multirow{2}{*}{\textbf{Backbone}} & \multirow{2}{*}{\textbf{\begin{tabular}[c]{@{}c@{}}FLOPs\\ (G)\end{tabular}}} & \multirow{2}{*}{\textbf{\begin{tabular}[c]{@{}c@{}}Params\\ (M)\end{tabular}}} & \multicolumn{6}{c|}{\textbf{Mask R-CNN 1x schedule}}                                              & \multicolumn{6}{c}{\textbf{Mask R-CNN 3x + MS schedule}}                                          \\ \cline{4-15} 
                                   &                                                                               &                                                                                    & \textbf{$AP^b$}  & \textbf{$AP^b_{50}$} & \textbf{$AP^b_{75}$} & \textbf{$AP^m$}  & \textbf{$AP^m_{50}$} & \textbf{$AP^m_{75}$} & \textbf{$AP^b$}  & \textbf{$AP^b_{50}$} & \textbf{$AP^b_{75}$} & \textbf{$AP^m$}  & \textbf{$AP^m_{50}$} & \textbf{$AP^m_{75}$} \\ \hline
ResNet-18~\cite{he2016deep}                          & 207                                                                           & 31                                                                                 & 34.0          & 54.0           & 36.7           & 31.2          & 51.0           & 32.7           & 36.9          & 57.1           & 40.0           & 33.6          & 53.9           & 35.7           \\
ResNet-50~\cite{he2016deep}                          & 260                                                                           & 44                                                                                 & 38.0          & 58.6           & 41.4           & 34.4          & 55.1           & 36.7           & 41.0          & 61.7           & 44.9           & 37.1          & 58.4           & 40.1           \\
ResNet-101~\cite{he2016deep}                         & 493                                                                           & 101                                                                                & 40.4          & 61.1           & 44.2           & 36.4          & 57.7           & 38.8           & 42.8          & 63.2           & 47.1           & 38.5          & 60.1           & 41.3           \\
PVTv1-T~\cite{wang2021pvtv1}                              & 208                                                                           & 33                                                                                 & 36.7          & 59.2           & 39.3           & 35.1          & 56.7           & 37.3           & 39.8          & 62.2           & 43.0           & 37.4          & 59.3           & 39.9           \\
PVTv1-S~\cite{wang2021pvtv1}                              & 245                                                                           & 44                                                                                 & 40.4          & 62.9           & 43.8           & 37.8          & 60.1           & 40.3           & 43.0          & 65.3           & 46.9           & 39.9          & 62.5           & 42.8           \\
PVTv2-B0~\cite{wang2022pvtv2}                              & 196                                                                           & 24                                                                                 & 38.2          & 60.5           & 40.7           & 36.2          & 57.8           & 38.6           & -          & -           & -           & -          & -           & -           \\
LightViT-T~\cite{huang2022lightvit}                         & 187                                                                           & 28                                                                                 & 37.8          & 60.7           & 40.4           & 35.9          & 57.8           & 38.0           & 41.5          & 64.4           & 45.1           & 38.4          & 61.2           & 40.8           \\
\textbf{DualToken-ViT-S (mix)}     & \textbf{191}                                                                  & \textbf{30}                                                                      & \textbf{41.1} & \textbf{63.5}  & \textbf{44.7}  & \textbf{38.1} & \textbf{60.5}  & \textbf{40.5}  & \textbf{43.7} & \textbf{65.8}  & \textbf{47.4}  & \textbf{39.9} & \textbf{62.7}  & \textbf{42.8}  \\ \hline
\end{tabular}
}
\label{table:maskrcnn}
\end{table*}

\begin{table}[htbp]
\centering
\caption{Object detection performance by RetinaNet on MS-COCO. All the models are pretrained on ImageNet-1K.}
\setlength\tabcolsep{2.3pt}
{\scriptsize
\begin{tabular}{c|cc|cccccc}
\hline
\textbf{Backbone} & \textbf{\begin{tabular}[c]{@{}c@{}}FLOPs\\ (G)\end{tabular}} & \textbf{\begin{tabular}[c]{@{}c@{}}Params\\ (M)\end{tabular}} & \textbf{$AP$} & \textbf{$AP_{50}$} & \textbf{$AP_{75}$} & \textbf{$AP_S$} & \textbf{$AP_M$} & \textbf{$AP_L$} \\ \hline
ResNet-18~\cite{he2016deep}         & 181                                                          & 21.3                                                          & 31.8        & 49.6          & 33.6          & 16.3         & 34.3         & 43.2         \\
ResNet-50~\cite{he2016deep}         & 239                                                          & 37.7                                                          & 36.3        & 55.3          & 38.6          & 19.3         & 40.0         & 48.8         \\
PVTv1-T~\cite{wang2021pvtv1}             & 221                                                          & 23.0                                                          & 36.7        & 56.9          & 38.9          & 22.6         & 38.8         & 50.0         \\
PVTv2-B0~\cite{wang2022pvtv2}             & 178                                                          & 13.0                                                          & 37.2        & 57.2          & 39.5          & 23.1         & 40.4         & 49.7         \\
ConT-M~\cite{yan2021contnet}            & 217                                                          & 27.0                                                          & 37.9        & 58.1          & 40.2          & 23.0         & 40.6         & 50.4         \\
MF-508M~\cite{chen2022mobileformer}           & 168                                                          & 17.9                                                          & 38.0        & 58.3          & 40.3          & 22.9         & 41.2         & 49.7         \\
\textbf{DualToken-ViT-S}      & \textbf{170}                                                          & \textbf{20.0}                                                          & \textbf{40.3}        & \textbf{61.2}          & \textbf{42.8}          & \textbf{25.5}         & \textbf{43.7}         & \textbf{55.2}         \\ \hline
\end{tabular}
}
\label{table:retinanet}
\end{table}

\section{Experiments}

\subsection{Image Classification}

\hspace*{\fill}

\noindent {\normalsize \bf Setting.} We perform image classification experiments on the ImageNet-1k~\cite{deng2009imagenet} dataset and validate the top-1 accuracy on its validation set. Our model is trained with 300 epochs and is based on 224$\times$224 resolution images. For the sake of fairness of the experiment, we try to choose models with this setup and do not use extra datasets and pre-trained models to compare with our model. We employ the AdamW~\cite{loshchilov2017decoupled} optimizer with betas (0.9, 0.999), weight decay 4e-2, learning rate 1e-3 and batch size 1024. And we use Cosine scheduler
with 20 warmup epoch. RandAugmentation
(RandAug (2, 9)), MixUp
(alpha is 0.2), CutMix
(alpha is 1.0), Random Erasing
(probability is 0.25), and drop path
(rate is 0.1) are also employed.

\noindent {\normalsize \bf Results.} We compare DualToken-ViT to other vision models on two scales of FLOPs, and the experimental results are shown in Table~\ref{table:cls}, where our model performs the best on both scales. For example, DualToken-ViT-S (mix) achieves 79.4\% accuracy at 1.0G FLOPs, exceeding the current SoTA model
LightViT-T~\cite{huang2022lightvit}.
And we improved the accuracy to 79.5\% after replacing MixMLP with normal MLP.

\begin{table*}[]
\caption{Semantic segmentation performance by DeepLabv3 and PSPNet on ADE20K dataset. All the models are pretrained on ImageNet-1K.}
\centering
\begin{tabular}{c|ccc|ccc}
\hline
\multirow{2}{*}{\textbf{Backbone}} & \multicolumn{3}{c|}{\textbf{DeepLabv3}}                       & \multicolumn{3}{c}{\textbf{PSPNet}}                           \\ \cline{2-7} 
                                   & \textbf{FLOPs (G)} & \textbf{Params (M)} & \textbf{mIoU (\%)} & \textbf{FLOPs (G)} & \textbf{Params (M)} & \textbf{mIoU (\%)} \\ \hline
MobileNetv2~\cite{sandler2018mobilenetv2}                        & 75.4               & 18.7                & 34.1               & 53.1               & 13.7                & 29.7               \\
MobileViTv2-1.0~\cite{mehta2022separable}                    & 56.4               & 13.4                & 37.0               & 40.3               & 9.4                 & 36.5               \\
\textbf{DualToken-ViT-S}           & \textbf{68.4}      & \textbf{26.3}       & \textbf{39.0}          & \textbf{58.3}      & \textbf{21.7}       & \textbf{38.8}          \\ \hline
\end{tabular}
\label{table:semanticseg}
\end{table*}

\subsection{Object Detection and Instance Segmentation}

\hspace*{\fill}

\noindent {\normalsize \bf Setting.} We perform experiments on the MS-COCO~\cite{lin2014microsoft} dataset and use RetinaNet~\cite{lin2017focal} and Mask R-CNN~\cite{he2017mask} architectures with FPN~\cite{lin2017feature} neck for a fair comparison. Since DualToken-ViT has only three stages, we modified the FPN neck using the same method as in LightViT~\cite{huang2022lightvit} to make our model compatible with these two detection architectures.
For the RetinaNet architecture, we employ the AdamW~\cite{loshchilov2017decoupled} optimizer for training, where betas (0.9, 0.999), weight decay 1e-4, learning rate 1e-4 and batch size 16. And we use the training schedule of 1$\times$ from the MMDetection library. For the Mask R-CNN architecture, we employ the AdamW optimizer for training, where betas (0.9, 0.999), weight decay 5e-2, learning rate 1e-4 and batch size 16. We use the 1$\times$ and 3$\times$ training schedules from the MMDetection library, respectively. We use all the standard metrics for object detection and instance segmentation of the MS-COCO dataset.

\noindent {\normalsize \bf Results.} We compare the performance of our model with other models on Mask R-CNN and RetinaNet architectures, and the experimental results are shown in Table~\ref{table:maskrcnn} and Table~\ref{table:retinanet}, respectively. Although our backbone has only three stages, DualToken-ViT-S without the maximum resolution stage still performs well in a model of the same FLOPs magnitude. In particular, in the experiments of Mask R-CNN architecture using the training schedule of 1$\times$, our backbone achieves 41.1\% $AP^b$ and 38.1\% $AP^m$ at 191G FLOPs, which far exceeds LightViT-T~\cite{huang2022lightvit} with similar FLOPs. This may be related to our position-aware global tokens, which we will explain in detail later.

\begin{table}[ht]
\setlength\tabcolsep{4.0pt}
\centering
\caption{Ablation study on the method of applying global tokens. Normal, Normal* and Position-aware represent the methods in Figure~\ref{fig:GlobalToken}, Figure~\ref{fig:PosNormGlobalToken} and Figure~\ref{fig:PosGlobalToken}, respectively.}
\begin{tabular}{c|ccc}
\hline
\textbf{Global Tokens}  & \textbf{FLOPs (G)} & \textbf{Params (M)} & \textbf{Top-1 (\%)} \\ \hline
Normal                  & 0.99               & 13.0                & 79.2                \\
Normal*                 & 0.98               & 13.4                & 79.0                \\
\textbf{Position-aware} & \textbf{1.05}      & \textbf{11.9}       & \textbf{79.5}       \\ \hline
\end{tabular}
\label{table:globaltokens}
\end{table}

\begin{table}[ht]
\caption{Ablation study on the number of tokens in position-aware global tokens.}
\centering
\begin{tabular}{c|ccc}
\hline
\textbf{Number} & \textbf{FLOPs (G)} & \textbf{Params (M)} & \textbf{Top-1 (\%)} \\ \hline
0               & 0.99              & 10.8                & 79.2              \\
3$\times$3               & 0.93              & 11.9              & 79.1              \\
4$\times$4              & 0.95              & 11.9              & 79.2              \\
5$\times$5     & 0.98              & 11.9              & 79.4       \\
6$\times$6              & 1.01              & 11.9              & 79.3              \\
\textbf{7$\times$7}     & \textbf{1.05}     & \textbf{11.9}     & \textbf{79.5}     \\
8$\times$8              & 1.10                & 11.9              & 79.3              \\ \hline
\end{tabular}
\label{table:numgtoken}
\end{table}

\begin{table}[ht]
\centering
\small
\caption{Ablation study on the method of local attention.}
\setlength\tabcolsep{3.8pt}
\begin{tabular}{c|ccc}
\hline
\textbf{Local Attention} & \textbf{FLOPs (G)} & \textbf{Params (M)} & \textbf{Top-1 (\%)} \\ \hline
Window Self-attention    & 0.92               & 10.8                & 78.6                \\
\textbf{Conv Encoder}    & \textbf{1.04}      & \textbf{11.4}       & \textbf{79.4}       \\ \hline
\end{tabular}
\label{table:localattention}
\end{table}

\begin{table}[ht]
\centering
\setlength\tabcolsep{4.3pt}
\caption{Ablation study on the step-wise downsampling part of the Position-aware Token Module.}
\begin{tabular}{c|ccc}
\hline
\textbf{Downsampling} & \textbf{FLOPs (G)} & \textbf{Params (M)} & \textbf{Top-1 (\%)} \\ \hline
one-step              & 1.01              & 11.3              & 79.2              \\
\textbf{step-wise}    & \textbf{1.04}     & \textbf{11.4}     & \textbf{79.4}     \\ \hline
\end{tabular}
\label{table:downsampling}
\end{table}

\subsection{Semantic Segmentation}

\hspace*{\fill}

\noindent {\normalsize \bf Setting.} We perform experiments on ADE20K~\cite{zhou2019semantic} dataset at 512$\times$512 resolution and use DeepLabv3~\cite{chen2017rethinking} and PSPNet~\cite{zhao2017pyramid} architectures for a fair comparison.
For training, we employ the AdamW~\cite{loshchilov2017decoupled} optimizer, where betas (0.9, 0.999), weight decay 1e-4, learning rate 2e-4 and batch size 32.

\noindent {\normalsize \bf Results.} We compare the performance of our model with other models on DeepLabv3 and PSPNet architectures, and the experimental results are shown in Table~\ref{table:semanticseg}. DualToken-ViT-S performs best among models of the same FLOPs magnitude on both architectures.

\subsection{Ablation Study}

\hspace*{\fill}

\noindent {\normalsize \bf MLPs.} We compare two MLPs performed on position-aware global tokens: normal MLP and MixMLP. The experimental results on DualToken-ViT-S are shown in Table~\ref{table:cls}. The normal MLP is 0.1\% more accurate than MixMLP, but it adds a little extra FLOPs and parameters. This is because MixMLP extracts information in the spatial dimension, it may damage some positional information on the position-aware global tokens.

\noindent {\normalsize \bf Different methods of applying global tokens.} We compare three methods of applying global tokens.
The method~\cite{huang2022lightvit, chen2022mobileformer, yao2022dualvit} in Figure~\ref{fig:GlobalToken} is the most common. Figure~\ref{fig:PosGlobalToken} shows our method that uses weighted summation to fuse $\bm{X}_{\text{ga}}$ and $\bm{G}$. Figure~\ref{fig:PosNormGlobalToken} combines the previous two methods, replacing the weighted summation based fusion in our method with the multi-head self-attention based fusion. We perform experiments on DualToken-ViT-S. In the implementation, because the complexity of the methods using multi-head self-attention based fusion is greater, we set the number of global tokens to 8, which is the same number as LightViT-T~\cite{huang2022lightvit}. The experimental results are shown in Table~\ref{table:globaltokens}, which show that our position-aware-based method performs the best and has 1.1M less parameters than the Normal method, with only 0.06G more FLOPs.
Since the other two methods employ multi-head self-attention based fusion that requires many parameters, whereas our method employs weighted summation based fusion, our method has the smallest parameters.
This demonstrates the superiority of position-aware global tokens.

\noindent {\normalsize \bf The number of tokens in position-aware global tokens.} We performed ablation study on the number of tokens in position-aware global tokens on ImageNet-1k~\cite{deng2009imagenet} dataset at 224$\times$224 resolution. In our model, the number of tokens in position-aware global tokens is set to 7$\times$7. In order to compare the impact of different numbers of tokens on our model,
we experiment with various settings for the number of tokens.
If the number of tokens is set to 0, then the position-aware global tokens are not used. Because the size of $\bm{X}_{\text{ga}}$ and the position-aware global tokens will not match when the number of tokens is not 7$\times$7, we will use interpolation for $\bm{X}_{\text{ga}}$ to make the size of the two match. The experimental results on DualToken-ViT-S are shown in Table~\ref{table:numgtoken}. The model with the number of tokens set to 7$\times$7 has the best performance due to the sufficient number of tokens and does not damage the information by the interpolation method. Compared to the 0 token setting, our setting is 0.3\% more accurate and will only increase by 0.06G FLOPs and 1.1M parameters, which demonstrates the effectiveness of our position-aware global tokens.

\begin{figure*}[ht]
\centering
    \includegraphics[width=\linewidth,trim=0 240 256 75,clip]{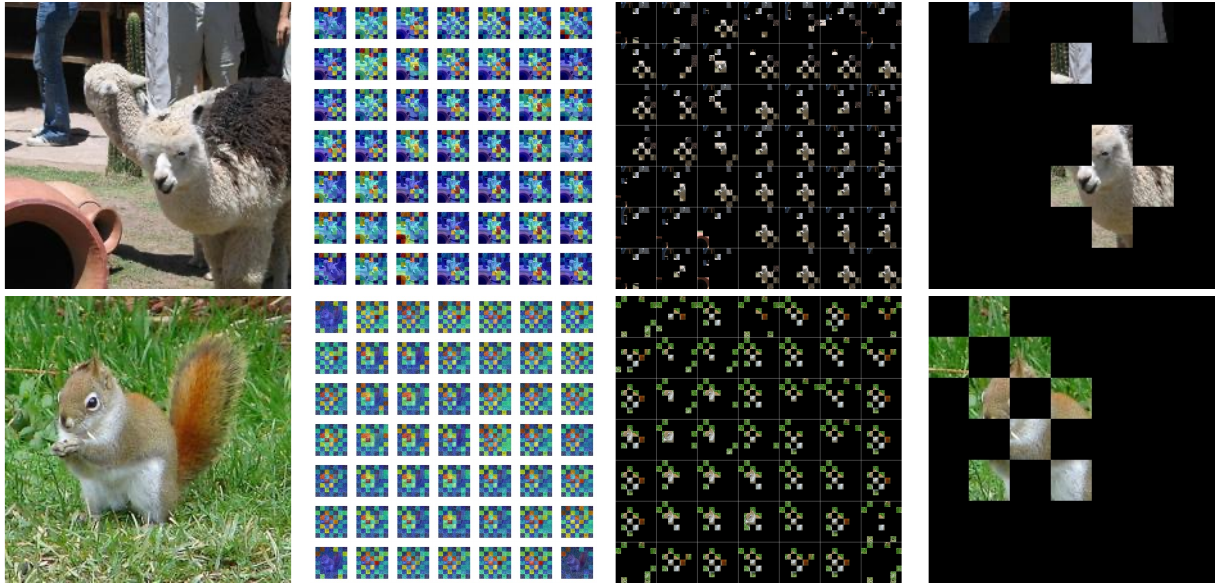}
    \caption{Visualization of the attention map of the Global Broadcast for the last block in our model. In each row, each subimage in the second image represents the correlation between this part of the first image and each token in the position-aware global tokens, and the third image shows the 8 tokens with the highest correlation in each subimage. The fourth image in each row represents the average of all subimages in the second image and shows the 8 tokens with the highest correlation.}
\label{fig:visual_all}
\end{figure*}

\noindent {\normalsize \bf Local attention.} We compare the role of Conv Encoder and window self-attention~\cite{liu2021swin} in our model.
And we set the window size of window self-attention to 7.
The experimental results on DualToken-ViT-S (mix) are shown in Table~\ref{table:localattention}. The model using Conv Encoder as local attention achieves better performance, with 0.8\% more accuracy than when using window self-attention, and the number of FLOPs and parameters does not increase very much. The performance of Conv Encoder is superior for two reasons. On the one hand, the convolution-based structure will be more advantageous than the transformer-based structure for light-weight models. On the other hand, window self-attention damages the position information in the position-aware global tokens. This is because the transformer-based structure does not have the inductive bias of locality. And in window self-attention, the features in the edge part of the window will be damaged due to the feature map being split into several small parts.

\noindent {\normalsize \bf Downsampling.} We perform ablation study on the step-wise downsampling part of the position-aware token module. For the setup of one-step downsampling, we directly downsample $\bm{X}_{\text{local}}$ to get the desired size, and then input it to the Global Aggregation. The experimental results on DualToken-ViT-S (mix) are shown in Table~\ref{table:downsampling}. Step-wise downsampling is 0.2\% more accurate than one-step downsampling, and FLOPs and parameters are only 0.03G and 0.1M more, respectively. The reason for this is that the method of step-wise can retain more information by convolution during the downsampling process.

\subsection{Visualization}

To get a more intuitive feel for the position information contained in position-aware global tokens, we visualize the attention map of the Global Broadcast for the last block in DualToken-ViT-S (mix),
and the results are shown in Figure~\ref{fig:visual_all}.
In each row, the second and third images show that the key tokens in the first image generate higher correlation with the corresponding tokens in the position-aware global tokens.
And in the second image in each row, the non-key tokens in the first image generate more uniform correlation with each part of the position-aware global tokens. The fourth image in each row shows that the overall position-aware global tokens have a higher correlation with the key tokens of the first image. These demonstrate that our position-aware global tokens contain position information.

\section{Conclusion}

In this paper, we propose a light-weight and efficient visual transformer model called DualToken-ViT. It achieves efficient attention structure by combining convolution-based local attention and self-attention-based global attention. We improve global tokens and propose position-aware global tokens that contain both global and position information. We demonstrate the effectiveness of our model on image classification, object detection and semantic segmentation tasks.

{\small
\bibliographystyle{siam}
\bibliography{egbib}
}

\end{document}